\pdfoutput=1

\documentclass[11pt]{article}
\usepackage{graphicx}

\usepackage[]{ACL2023}

\usepackage{times}
\usepackage{latexsym}
\usepackage{amsmath} 
\usepackage{bbold}
\usepackage{mathastext}
\usepackage{adjustbox}
\usepackage[T1]{fontenc}

\usepackage[utf8]{inputenc}

\usepackage{microtype}

\usepackage{inconsolata}
\usepackage{svg}
\usepackage{multirow}
\usepackage{hyperref}
\usepackage{booktabs}

\NewDocumentCommand{\heng}
{ mO{} }{\textcolor{red}{\textsuperscript{\textit{Heng}}\textsf{\textbf{\small[#1]}}}}

\title{LEMMA: Towards LVLM-Enhanced Multimodal Misinformation Detection with External Knowledge Augmentation}

\author{
    Keyang Xuan\thanks{~~Equal contribution.}, Li Yi\footnotemark[1], Fan Yang\footnotemark[1], Ruochen Wu, Yi R. Fung, Heng Ji\\ 
    University of Illinois Urbana-Champaign \\
    \texttt{\{keyangx3,liyi3,fy10,rw12,yifung2,hengji\}@illinois.edu}
}

\begin{document}
\maketitle
\begin{abstract}
The rise of multimodal misinformation on social platforms poses significant challenges for individuals and societies. Its increased credibility and broader impact make detection more complex, requiring robust reasoning across diverse media types and profound knowledge for accurate verification. The emergence of Large Vision Language Model (LVLM) offers a potential solution to this problem. Leveraging their proficiency in processing visual and textual information, LVLM demonstrates promising capabilities in recognizing complex information and exhibiting strong reasoning skills. We investigate the potential of LVLM on multimodal misinformation detection and find that even though LVLM has a superior performance compared to LLMs, its profound reasoning may present limited power with a lack of evidence. Based on these observations, we propose LEMMA: LVLM-Enhanced Multimodal Misinformation Detection with External Knowledge Augmentation. LEMMA leverages LVLM intuition and reasoning capabilities while augmenting them with external knowledge to enhance the accuracy of misinformation detection. Our external knowledge extraction module adopts multi-query generation and image source tracing to enhance the rigor and comprehensiveness of LVLM’s
reasoning. We observed that LEMMA improves the accuracy over the top baseline LVLM by \textbf{9\%} and \textbf{13\%} on \textit{Twitter} and \textit{Fakeddit} datasets respectively. \footnote{The code is available at \href{https://github.com/fan19-hub/LEMMA}{https://github.com/fan19-hub/LEMMA}}
\end{abstract}

\section{Introduction}
Multimodal misinformation, originating from the integration of multimedia on social platforms, raises significant concerns for individuals and societies. The contents of such misinformation can be readily consumed by the audience, often gaining a higher level of credibility and causing a border impact compared to textual misinformation \cite{doi:10.1080/10584609.2019.1674979, zannettou2018origins}. Unlike unimodal misinformation, detecting multimodal misinformation is more challenging, requiring robust reasoning to decipher cross-modal clues, coupled with the necessity for profound knowledge to verify the factuality of the essential information.

The rise of Large Language Models (LLMs) \cite{zhao2023survey} has significantly reshaped traditional NLP tasks, while recent efforts are leveraging LLMs to combat misinformation \cite{chen2023combating,hu2023bad}. However, these efforts are limited by LLMs' inability to process non-textual resources. Therefore, the recent emergence of Large Vision Language Models (LVLM) \cite{openai2023gpt4} provides a good opportunity to forward this line of research and here are several intuitions of adopting LVLM into combating multimodal misinformation: Firstly, the pretraining process with large-corpus provides LVLM with a profound understanding of real-world knowledge \cite{pmlr-v202-du23f} so that it has the potential to recognize complex information such as terms or entities appearing in the multimodality. Secondly, LVLM exhibits a strong reasoning capability through showcasing its remarkable performance on various tasks such as arithmetic reasoning \cite{amini-etal-2019-mathqa}, question answering \cite{kamalloo2023evaluating}, and symbolic reasoning \cite{wei2023chainofthought}. Thus, it has the potential to generate strong reasoning from multimodalities even in the zero-shot manner \cite{kojima2023large}. 
Moreover, LVLM presents a promising capability in incorporating external knowledge by utilizing retrieval-based tools, which has proved to be a beneficial functionality, particularly in tasks that demand fact-checking \cite{fatahi-bayat-etal-2023-fleek}.

Considering the aforementioned motivations, our primary objective is to investigate the following research questions: \textbf{Can LVLM effectively detect multimodal misinformation given their inherent capabilities?} We discover that LVLM can generally demonstrate satisfactory performance with its promising capability to process and reason about complex multimodal content. Despite these advances, current models still struggle when external contextual understanding is necessary for accurate misinformation detection. Traditional approaches to augmenting LLMs with external knowledge and up-to-date information, such as Retrieval-Augmented Generation (RAG) \cite{lewis2021retrievalaugmented}, often rely on directly generating queries from factual text. While effective for simple fact-checking, this method falls short in addressing the deceptive nature of multimodal misinformation. \cite{chen2023combating}. In addition, those methods usually can only capture semantic relevance and are unable to handle logical connections, resulting in information loss.

To bridge this gap, we introduce LEMMA: \textbf{L}VLM-\textbf{E}nhanced \textbf{M}ultimodal \textbf{M}isinformation Detection with External Knowledge \textbf{A}ugmentation. Unlike conventional methods which usually convert all modalities into textual information for analysis, LEMMA conducts parallel text and image searches to gather comprehensive evidence to enhance the quality of LVLM's reasoning. In addition, our approach utilizes a reasoning-aware multi-query generation that allows the model to evaluate the relevance of details within the broader misinformation context, thereby preventing over-focus on trivial details. What's more, we adopt a coarse to fine-grained distillation module that can effectively improve the quality of retrieval evidence. Our experiments show that LEMMA significantly improves accuracy over the top baseline LVLM by \textbf{9\%} and \textbf{13\%} on the \textit{Twitter} and \textit{Fakeddit} datasets. In summary, the major contributions of this paper are as follows:
\begin{itemize}
    \item We present a comprehensive empirical evaluation of LVLM capabilities on multimodal misinformation detection based on its inherited capability.
    \item We propose LEMMA, a simple yet effective LVLM-based approach that utilizes the benefits of LVLM intuition and reasoning capability with advanced, reasoning-based query generation and evidence filtering.
    \item We design an ad-hoc external knowledge extraction module that adopts multi-query generation and image source tracing to enhance the rigor and comprehensiveness of LVLM's reasoning.
\end{itemize}

\section{Related Work}
\subsection{Multimodal Misinformation Detection}
With the proliferation of multimedia resources, multimodal misinformation detection has gained increasing attention in recent years due to its potential threat to the dissemination of genuine information \cite{alam2022survey}. To identify multimodal misinformation, a traditional way is to evaluate the consistency between multimodality. To be specific, such evaluation can be realized by approaches such as multimodality feature representation learning \cite{10.1145/3219819.3219903, 10.1145/3292500.3330935, XUE2021102610}, using image captioning model \cite{zhou2020safe} and vision transformer \cite{DBLP:journals/corr/abs-2112-01131}. However, these methods usually rely on a deep learning-based model, which leads to the weakness of interpretability. To address this issue, \citet{liu-etal-2023-interpretable} tries to improve interpretability by integrating explainable logic clauses. In addition, \citet{fung-etal-2021-infosurgeon} proposes InfoSurgeon which attempts to solve this task by extracting fine-grained information in multimodality. However, this method presents limited precision and recall due to the limitation of automatic IE techniques. Furthermore, these methods suffer from the inherent limitations of the training process, which restrict their generalizability. Therefore, recently researchers have increasingly focused on leveraging LVLMs to tackle multimodal misinformation. After \citet{lyu2023gpt} illustrates LVLM's effectiveness in the task, key areas of this research extend to developing targeted solutions to combat specific types of multimodal misinformation \cite{qi2024sniffer}, addressing challenges related to domain shift \cite{liu2024fakenewsgpt4}, and enhancing interpretability \cite{wang2024mmidr}. These studies reflect LVLM as a promising solution to multimodal misinformation detection.

\subsection{Retrieval-Augmented Generation (RAG) for LLM/LVLM}
RAG is an advanced technique that combines the power of LLM/LVLM with information retrieval techniques. This method was originally designed to address the hallucination issue in text generation by LLMs \cite{NEURIPS2020_6b493230}. In addition, RAG approach is frequently applied in tasks requiring factual consistency, such as open-domain question answering \cite{zhu2021enhancing}, fact-checking \cite{maynez2020faithfulness} and code generation \cite{10.1145/3491101.3519665}, which demonstrates its promise. However, traditional RAG suffers from limitations such as static retrieval and lack of efficiency, which prompts researchers to develop more advanced versions to overcome these challenges. For example, \citet{Rackauckas_2024} demonstrates combining RAG and reciprocal rank fusion to improve comprehensiveness, \citet{mallen2023trust} proposes to evaluate query complexity based on entity frequency and
\citet{jeong2024adaptiverag} incorporates a question complexity classifier to adjust the external knowledge retrieval strategy for question answering. Meanwhile, \citet{merth2024superposition} introduces superposition prompting to process input documents in parallel and \cite{jin2024ragcache} improves the RAG efficiency through designing a multilevel dynamic caching system. Despite these advancements, the application of RAG in multimodal misinformation detection poses unique challenges. A critical aspect is the ability to discern and prioritize details that are crucial for identifying rumors while minimizing the retrieval of trivial details. To effectively address this requirement, our method incorporates a reasoning-based query generation approach, which guides the LVLM to focus on analyzing the most pertinent information first, thereby enabling targeted searches for external resources. 

\begin{figure*}[htb]
    \centering
    \includegraphics[width=1\textwidth]{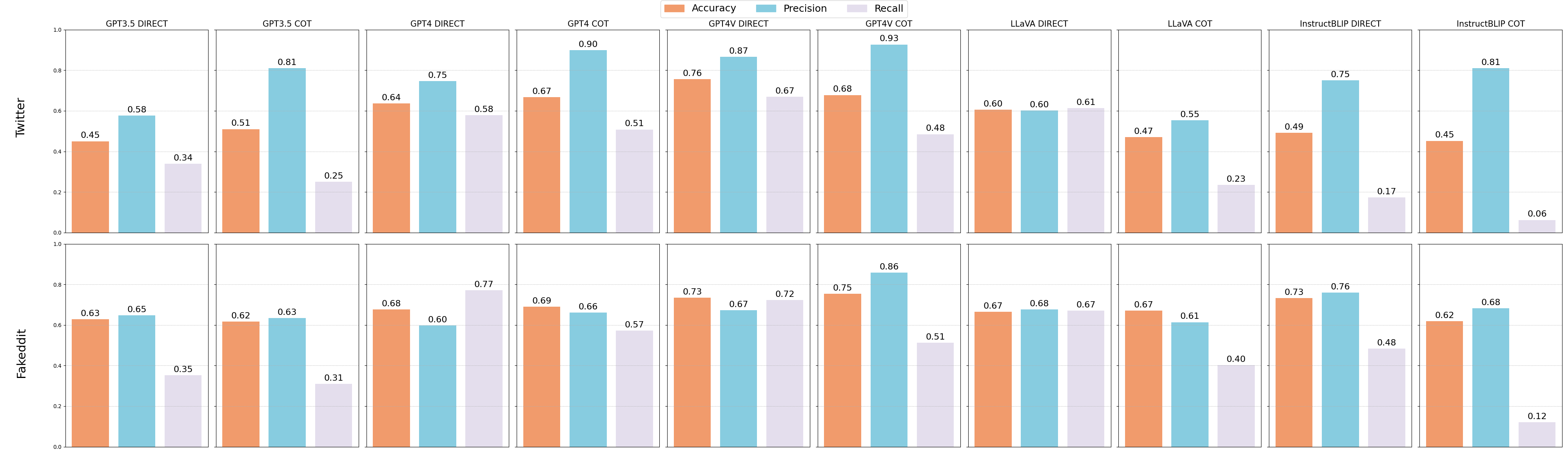}
    \caption{Comparison of performance metrics across various LVLMs/LLMs (GPT-3.5, GPT-4, GPT-4V, LLaVA, and InstructBLIP) and prompting methods (DIRECT and CoT) on two different datasets (\textit{Twitter} and \textit{Fakeddit}).}
    \label{fig: priliminary}
\end{figure*}

\section{Preliminary}
\subsection{Task Definition}
In this paper, our objective is to explore an LVLM-based solution for multimodal misinformation detection tasks. Given a post or news report which is formatted as an image-text pair $($\(\mathcal{I}\), \(\mathcal{T}\)$)$, we seek to classify it into a candidate label set \(\mathcal{Y}\) $ = \{ \text{NonMisinformation}, \text{Misinformation} \}$ based on two major criteria: \textbf{1)} whether there is an information inconsistency between \(\mathcal{I}\) and \(\mathcal{T}\) and \textbf{2)} whether there is a factuality issue in either \(\mathcal{I}\) or \(\mathcal{T}\). 

\subsection{Exploration}\label{sec: exploration}
\subsubsection{Evaluation Sets}
To assess the performance of LVLM on multimodal misinformation detection based on its inherent capability, we mainly evaluate its performance on two representative datasets in the field and the detailed stats for each dataset are presented in Appendix \ref{sec: Appendix A}.

\textit{\textbf{Twitter}} \cite{ma-etal-2017-detect} collects multimedia tweets from Twitter platform. The posts in the dataset contain textual tweets, image/video attachments, and additional social contextual information. For our task, we filtered out only image-text pairs as testing samples. 

\textit{\textbf{Fakeddit}} \cite{nakamura2019r} is designed for fine-grained fake news detection. The dataset is curated from multiple subreddits of the Reddit platform where each post includes textual sentences, images, and social context information. The 2-way categorization for this dataset establishes whether the news is real or false.

As LVLM doesn't necessitate a training phase, we leverage the testing sets directly from all evaluated datasets. Furthermore, we incorporate preprocessing by filtering out overly short tweets based on text length, as overly short texts are not able to provide sufficient information for inconsistency detection.

\subsubsection{Approaches}
We mainly exploit two fundamental prompting strategies for testing LVLM inherent capabilities on our task: \newline
\textbf{Direct}: In this method, we operate under the assumption that LVLM functions as an independent misinformation detector. Without applying any preprocessing techniques to image and text resources, we directly prompt LVLM to generate its prediction and then provide reasoning, relying solely on its internal knowledge. \newline
\textbf{Chain of Thought}: The Chain of Thought (CoT) mechanism  \cite{wei2023chainofthought} has demonstrated significant enhancement in the ability of LLMs to engage in complex reasoning tasks. Based on the Direct method, we further incorporate the phrase "\textit{Let's think step by step}" after the prompt. And LVLM is asked to first generate its reasoning and finally give out its prediction.

\subsubsection{Experiment Settings}
We take GPT-4V as a representative model to evaluate LVLM capability on multimodal misinformation detection. In our pursuit to understand the evolution of LVLMs, we also implement the aforementioned prompting approaches with GPT-3.5 and GPT-4. Since these models are not inherently multimodal, we conduct a preprocessing step by converting images into textual summaries to facilitate the input of multimodal content.
Additionally, to ensure a more comprehensive evaluation, we incorporate other two famous LVLMs, LLaVA \cite{liu2023visual} and InstructBLIP \cite{dai2023instructblip} into our experiments, which allows us to scrutinize how various LVLMs perform and to identify more general observations.

\begin{figure}[htb]
    \centering
    \includegraphics[width=0.4\textwidth]{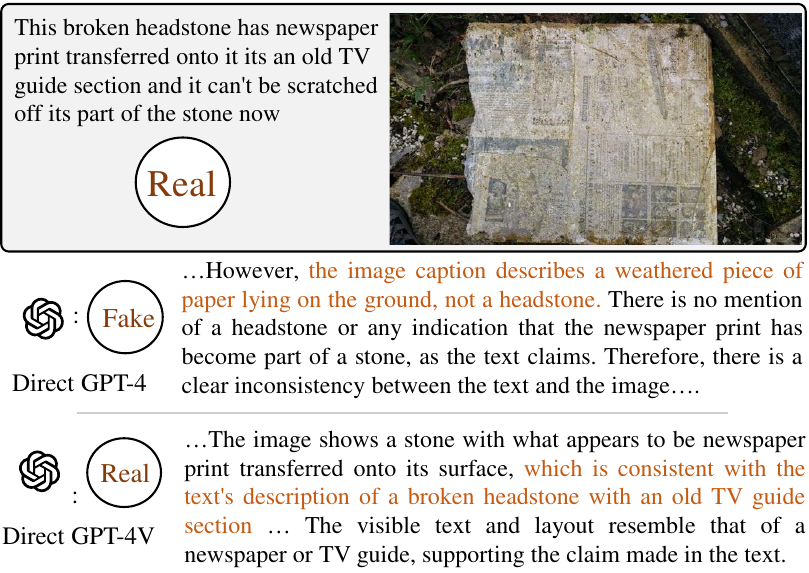}
    \caption{An example of a real \textit{Fakeddit} post where GPT-4V makes a correct prediction based on successfully extracting cross-modal alignment, while GPT-4 fails.}
    \label{fig: example1}
\end{figure}

\subsubsection{Observation on Preliminary Result}
Figure \ref{fig: priliminary} showcases the preliminary result of employing fundamental prompting strategies on two datasets using various LLMs/LVLMs. Upon scrutinizing the predictions and accompanying rationale, we deduce the following insights:

\begin{enumerate}
    \item \textbf{GPT-4V surpasses other LLMs/LVLMs in comprehending cross-modal interaction}: Across both datasets and prompting methods, GPT-4V demonstrates superior performance over other LLMs (like GPT-3.5 and GPT-4) and LVLMs (such as LLaVA and InstructBLIP). This superiority, when compared to LLMs, can be attributed primarily to its enhanced capability for multimodal understanding. For instance, Figure \ref{fig: example1} shows a real \textit{Fakeddit} post in which GPT-4V accurately extracts correlations between image and text. However, GPT4 struggles in extracting such correlation which eventually leads to a wrong decision. On the other hand, despite their pretraining for better multimodal capabilities, InstructBLIP and LLaVA tend to underperform due to their failure to follow instructions consistently and the mismatch between training corpus and specific task requirements, which eventually leads to the performance disparity in favor of GPT-4V.

    \item \textbf{In the absence of external evidence, reasoning-enhanced methods have very limited potential for performance improvement}: While CoT has already demonstrated superior performance in various tasks, its efficacy is limited in multimodal misinformation contexts when used with LVLMs. Specifically, while CoT may increase precision, it consistently yields lower recall compared to the Direct method, which suggests a tendency towards over-conservatism. Considering the importance of real-time information to misinformation detection, such conservative bias likely stems from the inherent limitations in reasoning without adequate supporting evidence, highlighting an essential trade-off between precision and recall in misinformation detection. For instance, Figure\ref{fig: example1} depicts a fabricated \text{Twitter} tweet that requires external evidence for an accurate decision. In such scenarios, CoT tends to guide LVLM towards a conservative stance.
\end{enumerate}

Based on these observations, although LVLM can achieve decent performance based on its inherent capability, it has limited power to make correct judgments when further evidence is necessary for the correct prediction. Therefore, with the insertion of external knowledge, LVLM is expected to achieve better performance.

\begin{figure}[htb]
    \centering
    \includegraphics[width=0.4\textwidth]{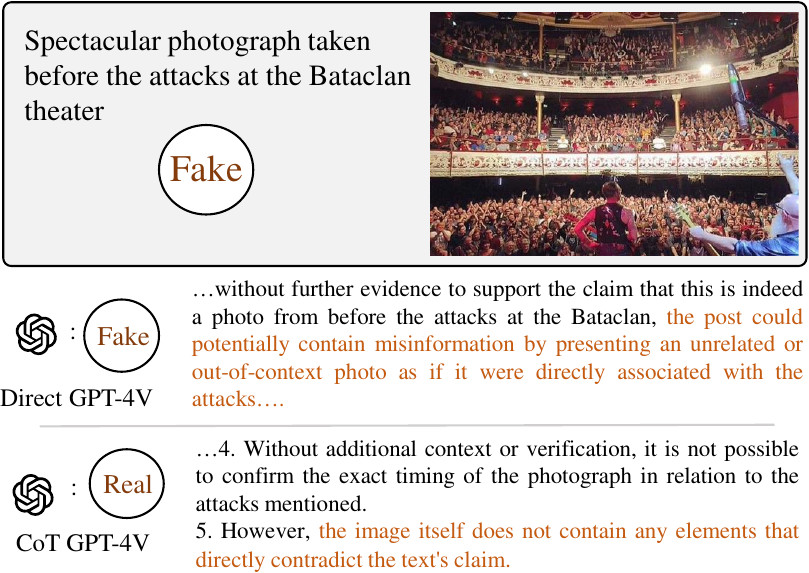}
    \caption{An example of a fabricated \textit{Twitter} tweet that shares subtle discrepancies in two modalities, misleading GPT-4V to answer "presence of misinformation"}
    \label{fig: example2}
\end{figure}

\section{Methodology}
\begin{figure*}[htb]
    \centering
    \includegraphics[width=1\textwidth]{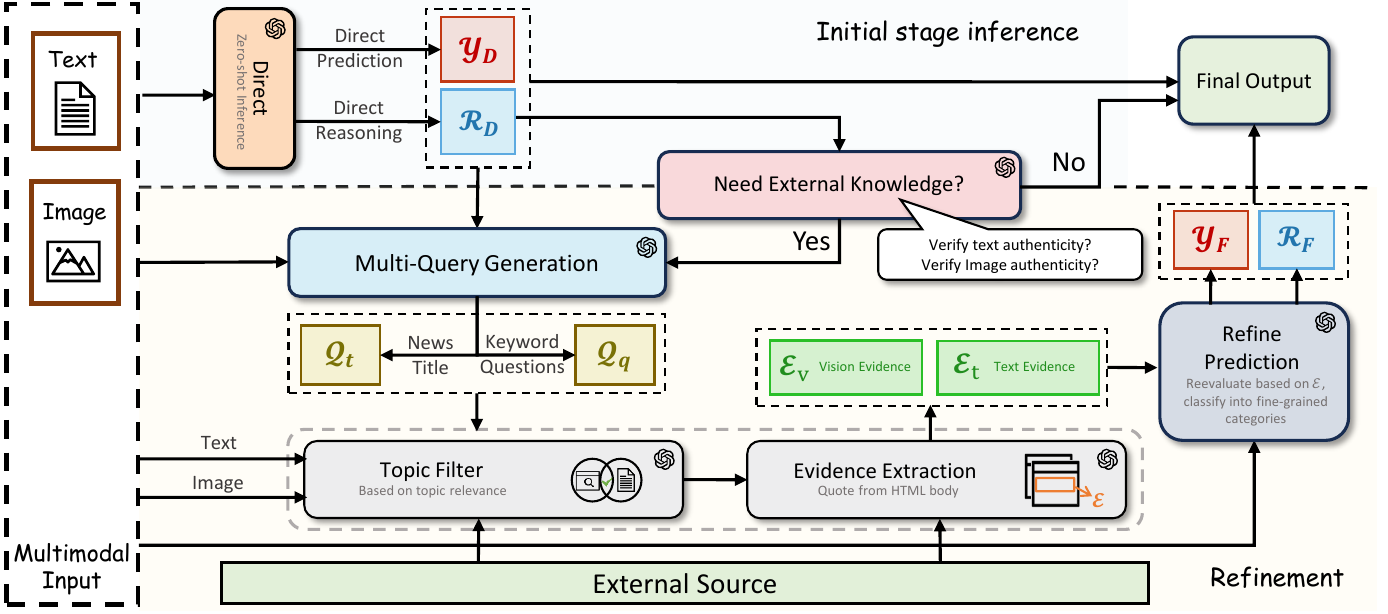}
    \caption{The pipeline of the proposed method (\textbf{LEMMA}). The process hinges on two key inputs: multimodal data and selectively filtered evidence gathered from external sources. Components marked with the OpenAI LOGO are developed using the LVLM (GPT-4V).}
    \label{fig: pipeline}
\end{figure*}

This section introduces the proposed \textbf{L}VLM-\textbf{E}nhanced \textbf{M}ulimodal \textbf{M}isinformation Detection with External Knowledge \textbf{A}ugmentation (\textbf{LEMMA}). The pipeline of LEMMA is illustrated in Figure \ref{fig: pipeline}. We first delve into the initial stage inference in Section \ref{sec: first stage}. Subsequently, we elucidate how we generate reasoning-aware queries to retrieve relevant multimodal evidence from the Internet in Section \ref{sec: question gen}. Additionally, we present the methodology for filtering qualified evidence from search results in Section \ref{sec: tl method}. Finally, we demonstrate how LEMMA utilizes additional references to refine its final prediction in Section \ref{sec: lemma method}. The detailed prompt design for each module is shown in Appendix \ref{sec: LEMMA prompt}.

\subsection{Initial Stage Inference}\label{sec: first stage}
In the initial phase, LVLM assesses whether posts inherently contain misinformation based on observed cross-modal inconsistencies, and determines whether external information is necessary to make a final judgment. Upon receiving an image-text pair $($\(\mathcal{I}\), \(\mathcal{T}\)$)$, LVLM generates an initial prediction $\mathcal{Y}_D$ and accompanying rationale $\mathcal{R}_D$ which includes the assessment of consistency level between \(\mathcal{I}\) and \(\mathcal{T}\). Subsequently, leveraging reasoning $\mathcal{R}_D$, LVLM is able to autonomously evaluate the necessity for external knowledge based on whether the within-context information is sufficient to conclude the judgment and whether any contents need to be verified. Following this evaluation, LVLM will finalize its decision as the direct prediction if the current information is deemed sufficiently comprehensive. Otherwise, LVLM proceeds to extract external evidence for further analysis to avoid an overly conservative bias. Furthermore, if LVLM thinks the external knowledge is still insufficient for judgment, it will classify this post as "Unverified" in the refined prediction phrase and choose direct prediction instead as the final output. More details in Section  \ref{sec: lemma method}. 

\subsection{Multimodal Retrieval} \label{sec: question gen}
In addressing the challenge of potentially conservative bias due to insufficient evidence, we proposed a multimodal retrieval framework that combines reasoning-aware multi-query-based text retrieval and image context retrieval.

\subsubsection{Reasoning Aware Multi-Query Retrieval} 
Traditional retrieval methods often directly use original posts for query construction, leading to potential losses in semantic integrity and difficulties in matching dispersed information  \cite{mallen2023trust, shi2023large}. To address these, we employ LVLM to generate multi-faceted queries based on the direct reasoning $\mathcal{R}_D$. Specifically, LVLM receives the image-text pair $($\(\mathcal{I}\), \(\mathcal{T}\)$)$, along with the initial prediction $\mathcal{Y}_D$ and the reasoning $\mathcal{R}_D$ generated during the initial stage inference. LVLM first synthesizes a concise title $\mathcal{Q}_t$ for the post, where a "fake news" prefix is added to increase the likelihood of retrieving content that directly refutes the claims made in $\mathcal{T}$. Then, it reviews direct reasoning $\mathcal{R}_D$ that identifies the key discrepancies and statements that would suggest potential misinformation and raises several questions $\mathcal{Q}_q$ to verify them. This ensures that the system prioritizes areas most susceptible to misinformation. 

The combined query set $(\mathcal{Q}_t, \mathcal{Q}_q)$, is used to search via the DuckDuckGo Search API \cite{DuckDuckGo}, aiming to retrieve highly relevant documents set $\mathcal{D}$, each annotated with a web title and brief description.

\begin{figure}[htb]
    \centering
    \includegraphics[width=0.5\textwidth]{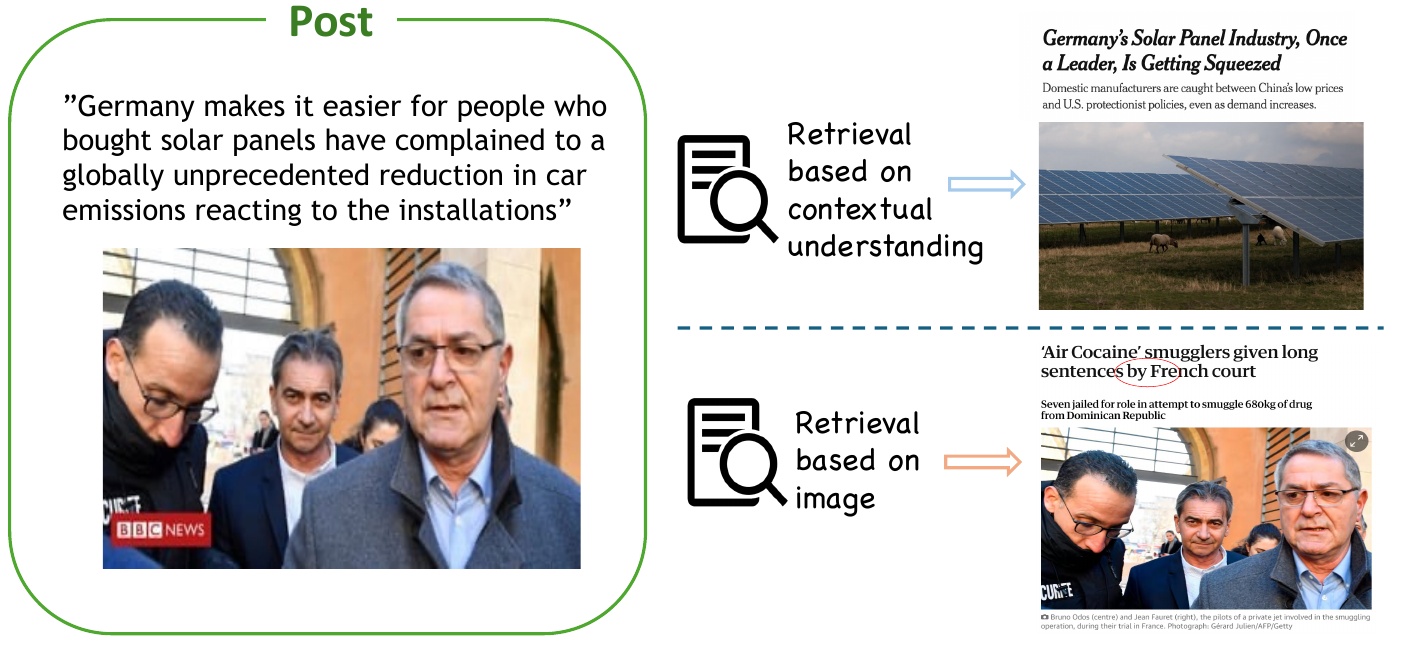}
    \caption{A fake news example that image retrieval exposes as a reused promotional image}
    \label{fig: img_retri}
\end{figure}

\subsubsection{Image Context Retrieval} 
Traditional retrieval methods often transform multimodal content into textual representations to facilitate analysis, focusing mainly on contextual comprehension. However, as depicted in Figure \ref{fig: img_retri}, this approach may overlook essential aspects of misinformation. For example, an image purported to show recent environmental benefits from new solar panels in Germany is actually an old promotional image. To address such discrepancies, image context retrieval technique provides a substantial improvement. By tracing the origin of an image, visual search adds a layer of context that significantly enhances the accuracy of misinformation detection.

To implement the image context retrieval, we utilize the Google search engine to trace the sources of image \(\mathcal{I}\) and the exact match technique to pinpoint the sources which contain the pictures that are identical or highly similar to \(\mathcal{I}\). Eventually, a list of web page's title is returned as evidence $\mathcal{E}_v$, which can offer a more accurate estimation of the image's context for later evaluation.

\subsection{Resource Distillation}\label{sec: tl method}
To address the challenge of off-topic or irrelevant information retrieved by search engines, we employ a resource distillation process, refining the traditional chunking technique based on the vector space model which lacks awareness of logical text connections  \cite{NEURIPS2020_6b493230}. We adopt a coarse to fine-grained distillation approach, similar to LongLLMLingua \cite{jiang2023longllmlingua}

\subsubsection{Topic Filtering}
Initially, the top k relevant resources form a root document set \(\mathcal{D}\). The LVLM then evaluates the topic relevance level of each document in \(\mathcal{D}\) based on query $(\mathcal{Q}_t, \mathcal{Q}_q)$ and original context \(\mathcal{I}\). Eventually, a further refined set \(\mathcal{D'}\) is returned, containing only documents that are highly relevant to the post's content. To ensure efficiency, we ask LLM to process a batch of resources in one request.


\subsubsection{Evidence Extraction}
For each document in $\mathcal{D'}$, we extract the main content along with the publication date. Subsequently, the LLM identifies key segments $\mathcal{S}_i$ that either support or refute the original post $\mathcal{T}$. The LLM is instructed to extract these segments directly from the HTML body of the document, ensuring they are succinct yet comprehensive, capturing all relevant information. These segments, along with the document's web title and publication date, are then compiled into an evidence entry, formatted as a triplet. The aggregated evidence, $\mathcal{E}_t$, is a collection of these triplets, forming a structured dataset ready for analysis.

\begin{table*}[htb]
  \centering
  \resizebox{0.9\textwidth}{!}{
  \resizebox{!}{0.8\textheight}{
  \begin{tabular}{clccccccc}
    \hline\hline
    \multirow{2}{*}{\textbf{Dataset}} & \multirow{2}{*}{\textbf{Method}} & \multirow{2}{*}{\textbf{Accuracy}} & \multicolumn{3}{c}{\textbf{Rumor}} & \multicolumn{3}{c}{\textbf{Non-Rumor}} \\
    \cmidrule(lr){4-6} \cmidrule(lr){7-9}
    & & & \textbf{Precision} & \textbf{Recall} & \textbf{F1} & \textbf{Precision} & \textbf{Recall} & \textbf{F1} \\
    \hline
    \multirow{12}{*}{\textit{Twitter}} &
    {\small Direct (LLaVA)} & 0.605 & 0.688 & 0.590 & 0.635 & 0.522 & 0.626 & 0.569 \\
     & {\small CoT (LLaVA)} & 0.468 & 0.563 & 0.231 & 0.635 & 0.441 & 0.765 & 0.560 \\
     & {\small Direct (InstructBLIP)} & 0.494 & 0.751 & 0.171 & 0.277 & 0.443 & 0.902 & 0.599 \\
     & {\small CoT (InstructBLIP)} & 0.455 & 0.813 & 0.067 & 0.112 & 0.428 & 0.921 & 0.596 \\
     & {\small Direct (GPT-4)} & 0.637 & 0.747 & 0.578 & 0.651 & 0.529 & 0.421 & 0.469 \\
     & {\small CoT (GPT-4)} & 0.667 & 0.899 & 0.508 & 0.649 & 0.545 & 0.911 & 0.682 \\
     & {\small FacTool (GPT-4)} &  0.548 & 0.585 & \textbf{0.857} & 0.696 & 0.273 & 0.082 & 0.125 \\
     & {\small Direct (GPT-4V)} & 0.757 & 0.866 & 0.670 & 0.756 & 0.673 & 0.867 & 0.758 \\
     & {\small CoT (GPT-4V)} & 0.678  & 0.927 & 0.485 & 0.637 & 0.567 & \underline{0.946}& 0.709  \\
     \cline{2-9}
     & {\small \textbf{LEMMA}} & \textbf{0.824} & \underline{0.943}& \underline{0.741} & \textbf{0.830} & \textbf{0.721} & 0.937& \textbf{0.816} \\
     
     & {\small \ w/o \textit{initial-stage infer}} & \underline{0.809}& 0.932& 0.736& \underline{0.823}& \underline{0.699}& 0.919& \underline{0.794}\\
     & {\small \ w/o \textit{visual retrieval}} & 0.781& \bf{0.953}& 0.672& 0.788& 0.652& \bf{0.949}& 0.773\\
    \hline
    \multirow{12}{*}{\textit{Fakeddit}} & {\small Direct (LLaVA)} & 0.663 & 0.588 & \underline{0.797} & 0.677 & 0.777 & 0.558 & 0.649 \\
    & {\small CoT (LLaVA)} & 0.673 & 0.612 & 0.400 & 0.484 & 0.694 & 0.843 & 0.761 \\
     & {\small Direct (InstructBLIP)} & 0.726 & 0.760 & 0.489 & 0.595 & 0.715 & 0.892 & 0.793 \\
     & {\small CoT (InstructBLIP)} & 0.610 & 0.685 & 0.190 & 0.202 & 0.604 & 0.901 & 0.742 \\
    & {\small Direct (GPT-4)} & 0.677 & 0.598 & 0.771 & 0.674 & 0.776 & 0.606 & 0.680 \\
     & {\small CoT (GPT-4)} & 0.691 & 0.662 & 0.573 & 0.614 & 0.708 & 0.779 & 0.742 \\
     & {\small FacTool (GPT-4)} & 0.506 & 0.476 & \textbf{0.834} & 0.606 & 0.624 & 0.232 & 0.339 \\
     & {\small Direct (GPT-4V)} & 0.734 & 0.673 & 0.723 & 0.697 & 0.771 & 0.742 & 0.764 \\
     & {\small CoT (GPT-4V)} & 0.754 & \underline{0.858} & 0.513 & 0.642 & 0.720 & \bf{0.937}& 0.814 \\
      \cline{2-9}
     & {\small \textbf{LEMMA}} & \textbf{0.828} & \textbf{0.881} & 0.706 & \textbf{0.784} & \bf{0.800}& \underline{0.925}& \textbf{0.857} \\
    
     & {\small \ w/o \textit{initial-stage infer}} & \underline{0.803}& 0.857 & 0.692 & \underline{0.766}& \underline{0.786}& 0.891 & 0.830 \\
     & {\small \ w/o \textit{visual retrieval}} & 0.792& 0.818& 0.675& 0.740& 0.778& 0.883& \underline{0.854} \\
     \hline\hline
  \end{tabular}
  }
  }
  \caption{Performance comparison of baseline methods and LEMMA on \textit{Twitter} and \textit{Fakeddit} dataset. We show the result of eight different baseline methods. Additionally, we present the results of two ablation studies: one without initial-stage inference, and the other without resource distillation and evidence extraction. The best two results are \textbf{bolded} and \underline{underlined}.}
  \label{table:comparison}
\end{table*}

\subsection{Refined Prediction}\label{sec: lemma method}
With the set of extracted evidence $(\mathcal{E}_t, \mathcal{E}_v)$ collected from external sources, the model gains a more comprehensive understanding of the multimodal content, enabling it to make a more accurate prediction. In detail, the image-text pair $($\(\mathcal{I}\), \(\mathcal{T}\)$)$ is re-introduced to the LVLM, accompanied with the evidence set $(\mathcal{E}_t, \mathcal{E}_v)$. LVLM is tasked with reevaluating its decision in light of the extracted evidence. Inspired by the fine-grained definition of multimodal misinformation \cite{nakamura2019r}, LVLM is asked to categorize the post into one of six categories: 1) True, 2) Satire, 3) Misleading Content, 4) False Connection, 5) Manipulated Content, or 6) Unverified Content. Categories 2 through 6 correspond to different types of misinformation, while Category 1 indicates real news. LVLM retains its inference from the initial stage if it classifies the post as Category 6, prioritizing conservatism over a potentially risky choice.

\section{Experiments}
\subsection{Experiment Settings}
We evaluate LEMMA by comparing it with the following baseline models and methods:
\textbf{1) LLaVA:} We evaluate LLaVA-1.5-13B \cite{liu2023visual}, which is a state-of-the-art LVLM based on vision instruction tuning, by employing the Direct approach. \textbf{2) InstructBLIP:} We evaluate the InstructBLIP \cite{dai2023instructblip}, which is a multimodal transformer designed to perform image-text tasks by leveraging instruction-based finetuning. \textbf{3) GPT-4 with Image Summarization:} We evaluate the effectiveness of the fundamental GPT-4 model (without visual understanding). To provide visual context, we construct a GPT4-V-based Image Summarization module, which generates comprehensive textual descriptions corresponding to images. As elaborated in Section \ref{sec: exploration}, we employ both the Direct and CoT approaches within this experimental framework. \textbf{4) GPT-4 with Factool:}We evaluated FacTool \cite{chern2023factool} with GPT-4 and image summarization as its foundation. Factool is an LLM-based framework that can detect factual inaccuracies within texts. Similar to LEMMA, FacTool incorporates query generation and evidence retrieval to verify claims. However, its methodology is specifically tailored for AI-generated text and does not take LVLM as a backbone. 
\textbf{5) GPT-4V:} We evaluate GPT-4V, also employing the Direct and CoT approaches.\\
\textbf{Datasets:} We evaluate LEMMA and all the baselines on the \textit{Twitter} and the \textit{Fakeddit} datasets, as introduced in \ref{sec: exploration}.

\subsection{Performance Comparison} \label{sec: performance}
The results presented in Table \ref{table:comparison} demonstrate that our proposed LEMMA framework consistently surpasses baseline models on the \textit{Twitter} and \textit{Fakeddit} datasets in terms of both Accuracy and F1 Score. Specifically, LEMMA shows an improvement of approximately 6.7\% in accuracy on \textit{Twitter} and a notable 7.4\% increase on \textit{Fakeddit} when compared to the best-performing baseline. Compared to FacTool which suffers from an overemphasis on trivial details that makes its predictions overly sensitive, our approach excels in balancing precision and recall, achieving high scores in both metrics. This suggests that LEMMA is effective in minimizing both false positives and false negatives, enhancing the overall quality of its predictions. Additionally, LEMMA demonstrates robust performance across different datasets, confirming its reliability and effectiveness in diverse contexts, essential for practical applications.

\begin{figure}[htb]
    \centering
    \includegraphics[width=0.46\textwidth]{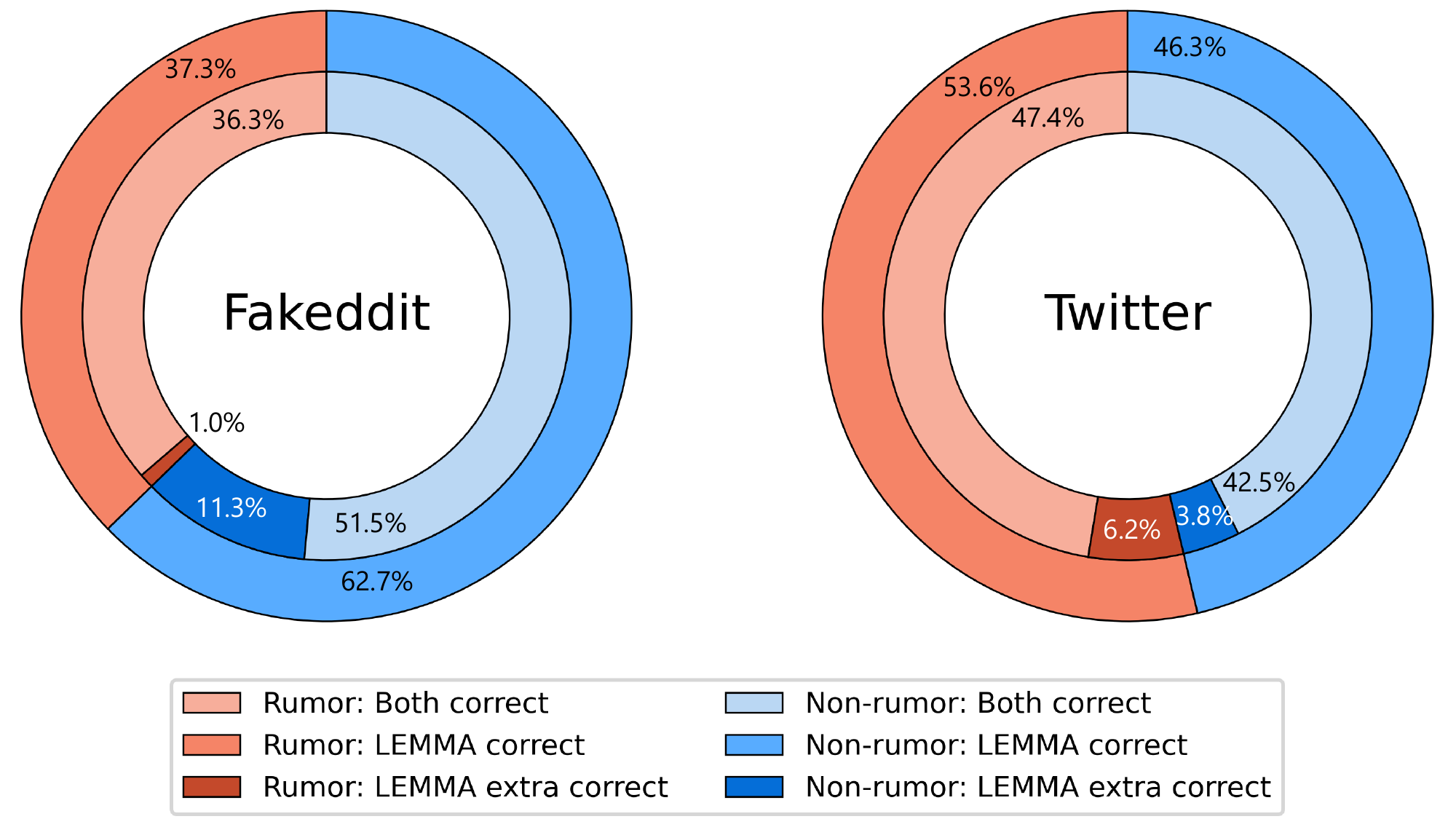}
    \caption{Comparison of the distribution of correct predictions between LEMMA and baseline (GPT-4V).}
    \label{fig: pie}
\end{figure}


\subsection{Ablation Study}
We conduct an ablation study on two modules in LEMMA, with the results shown in Table \ref{table:comparison}. (\textit{i}) \textit{Initial-stage inference.} We test bypassing LVLM's self-evaluation of external evidence necessity, forcing it to search for external evidence for all posts. This led to a 1.5\% lower accuracy on Twitter and a 2.5\% decrease on Fakeddit compared to the original version. We hypothesize that this is because LEMMA may be overly sensitive to the subtle differences between the external evidence and the original post. (\textit{ii}) \textit{Visual Retrieval.} We also implement a version without visual context retrieval, resulting in a 0.8\% drop in accuracy on Twitter and a 0.6\% drop on Fakeddit, suggesting that the image sources provide valuable context, informing LVLM of the true significance behind the visual input, thereby enhancing the overall reasoning quality.


\subsection{Result Analysis}
We conduct a statistical analysis to compare the accuracy distribution between LEMMA and Direct (GPT-4V). From Figure \ref{fig: pie}, we have the following observations: First, we observe that LEMMA accurately replicates over 98\% of Direct (GPT-4V) correct predictions in \textit{Fakeddit}, while in \textit{Twitter}, this figure stands at over 96\%. This suggests that LEMMA maintains an advantage in retaining the inherent capabilities of GPT-4V. Furthermore, in \textit{Fakeddit} and \textit{Twitter}, LEMMA exhibits approximately 13\% and 9\% additional gains relative to Direct (GPT-4V). Such performance advantages can be attributed to external knowledge providing LEMMA with more evidence favorable for inference, thereby making its reasoning performance more robust. 


\section{Conclusion}
In this study, we investigated the capability of LVLMs in multimodal misinformation detection and discovered the significant importance of providing external information to enhance LVLM performance. Then we proposed LEMMA, a framework designed to enhance LVLMs by utilizing a reasoning-aware query set for effective multimodal retrieval and by integrating external knowledge sources. Our experiments on the Twitter and Fakeddit datasets demonstrated that LEMMA significantly outperforms the top baseline LVLM, achieving accuracy improvements of \textbf{9\%} and \textbf{13\%}, respectively. While there is room for further refinement of knowledge source interfaces and filtering, we believe LEMMA is an extensible approach applicable to interpretability-critical reasoning tasks at the intersection of vision, language, and verification. 

\section{Limitations}
We recognize several limitations. 1) Due to the integration of external knowledge sources and multiple LVLM-based modules, the LEMMA framework may suffer from increased computational complexity and latency. This setup can hinder its scalability and efficiency, particularly in real-time environments where rapid processing is crucial. 2) Our study did not thoroughly examine LEMMA's sensitivity to different prompts. Given the constraints of our study, we defer the exploration of prompt sensitivity to future experiments. 3) The Evaluation datasets are limited to short social media posts due to dataset availability constraints, leaving LEMMA's performance on longer texts untested.

\section{Ethics Statement}
We acknowledge that our work is aligned with the ACL Code of the Ethics \footnote{https://www.aclweb.org/portal/content/
acl-code-ethics}
and will not raise ethical
concerns. We do not use sensitive datasets/models
that may cause any potential issues/risks.

\bibliography{anthology,custom}
\bibliographystyle{acl_natbib}

\newpage
\onecolumn
\appendix
\section{Dataset Statistics} \label{sec: Appendix A}
Table \ref{tab:dataset_stats} shows the detailed statistics of two datasets for the testing. We filter the original test sets of the two datasets to exclude overly short texts because they often lack sufficient contextual details. This means overly short texts are not good test cases to determine LVLM's capability for multimodal misinformation detection.

\begin{table*}[ht]
\centering
\begin{tabular}{|c|c|c|c|}
\hline
\textbf{Dataset} & \textbf{Num Rumor} & \textbf{Num Non-rumor} & \textbf{Language Distribution} \\ \hline
Twitter & 448 & 321 & English 78\%, French 9\%, Spanish 4\%, Other 9\% \\ \hline
Fakeddit & 342 & 464 & English 99\%, Other 1\% \\ \hline
\end{tabular}
\caption{Dataset Statistics}
\label{tab:dataset_stats}
\end{table*}

\section{Experiment Prompt Template} \label{sec: Appendix B}
This section shows the templates for each prompting method that we have examined in section \ref{sec: performance}:

\subsection{Direct Prompting} \label{sec:direct prompt}
\begin{figure*}[htb]
    \centering
    \includegraphics[width=\textwidth]{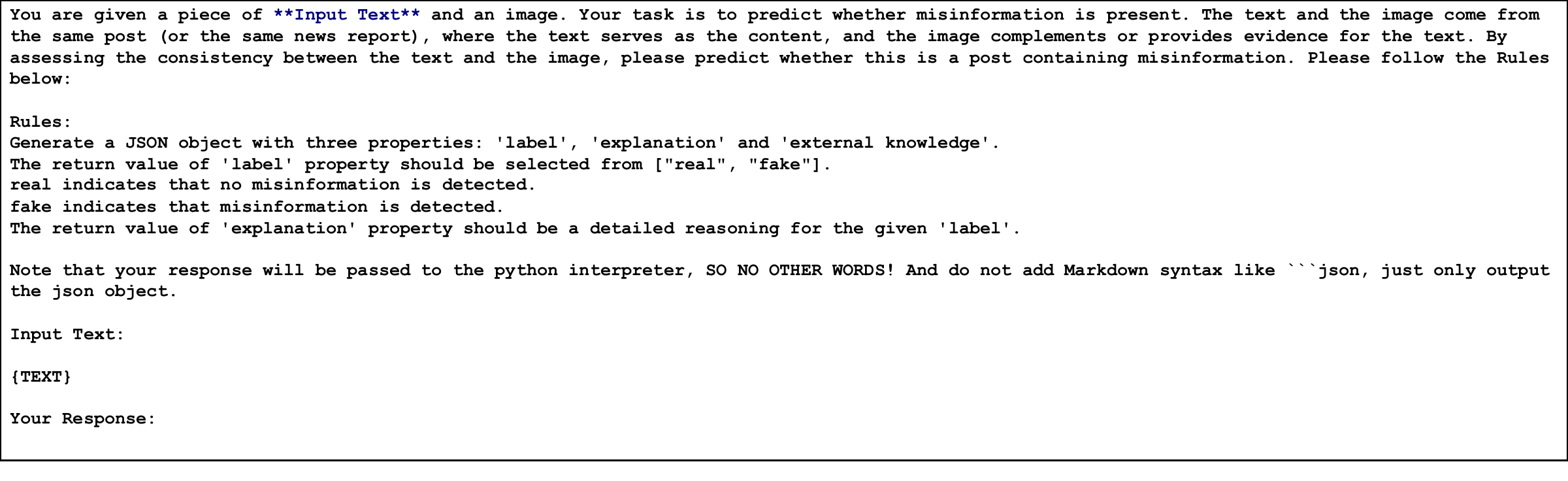}
    \caption{Prompt Template for Direct Approach}
    \label{fig: direct template}
\end{figure*}

As shown in Figure \ref{fig: direct template}, this method involves directly prompting the model with a text-image pair to determine the presence of misinformation. It must be noted that the rules we provided are designed to streamline the assessment process, ensuring that the model controls the output format while focusing on key indicators of misinformation.

\subsection{CoT Prompting}
\begin{figure*}[htb]
    \centering
    \includegraphics[width=\textwidth]{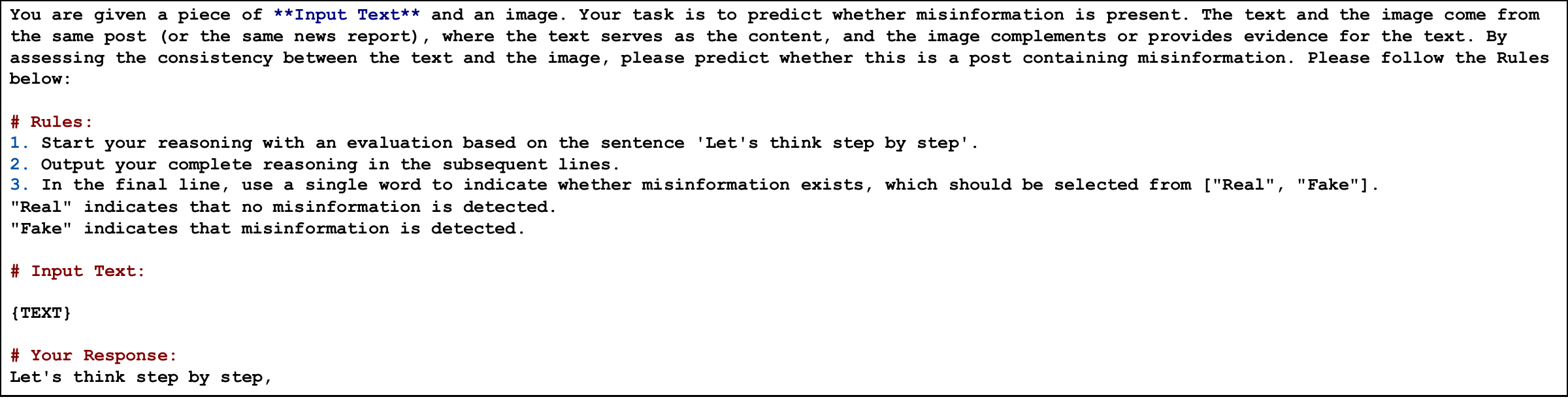}
    \caption{Prompt Template for CoT Approach}
    \label{fig: CoT prompt}
\end{figure*}

As illustrated in Figure \ref{fig: CoT prompt}, CoT extends the assessment process by incorporating a more explicit and detailed reasoning pathway. Similar to the \ref{sec:direct prompt}, the output format remains controlled.

\subsection{LEMMA Prompting} \label{sec: LEMMA prompt}
\subsubsection{Initial Stage Inference} \label{sec:initial stage prompt}
\begin{figure*}[htb]
    \centering
    \includegraphics[width=0.95\textwidth]{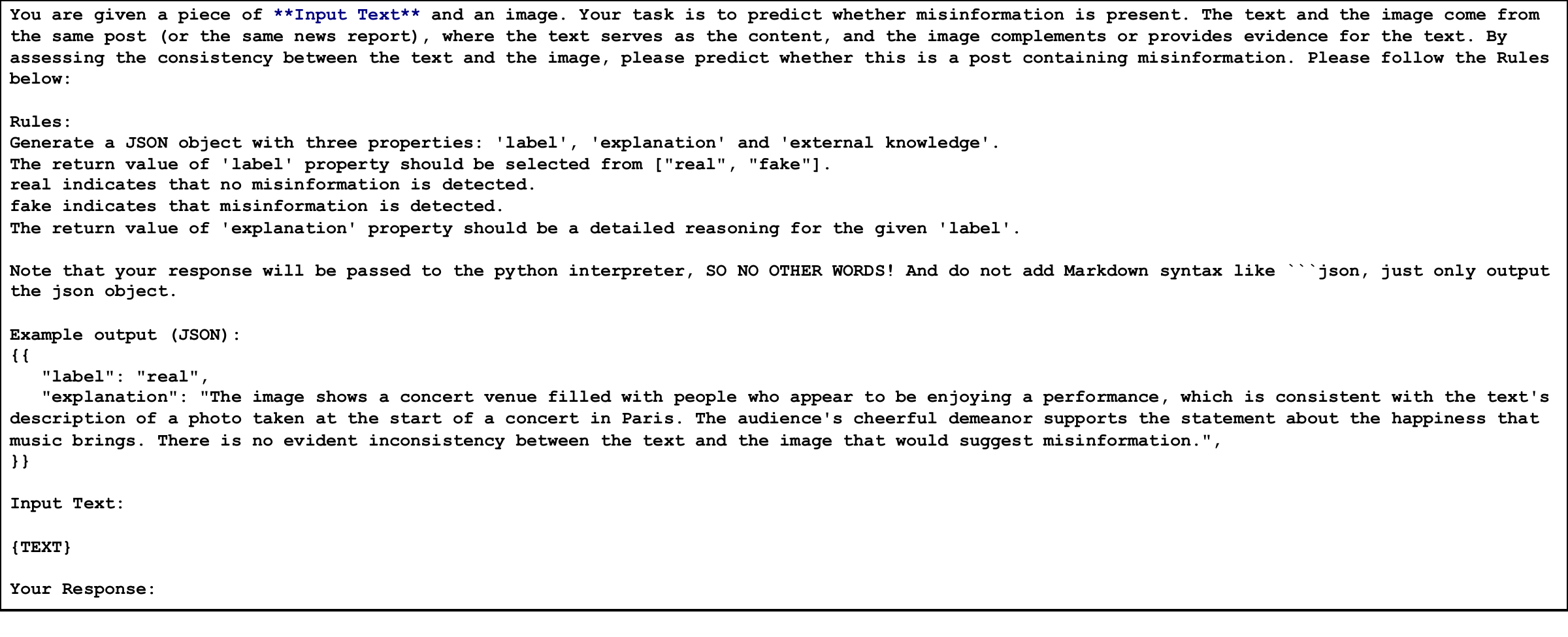}
    \caption{Instruction for initial stage inference}
    \label{fig: inference stage}
\end{figure*}

Shown in Figure \ref{fig: inference stage}, The prompt for this stage resembles the one from \ref{sec:direct prompt}, with the addition of an example to preserve the output format. This is crucial for deriving the reasoning needed for subsequent steps.

\begin{figure*}[htb]
    \centering
    \includegraphics[width=0.95\textwidth]{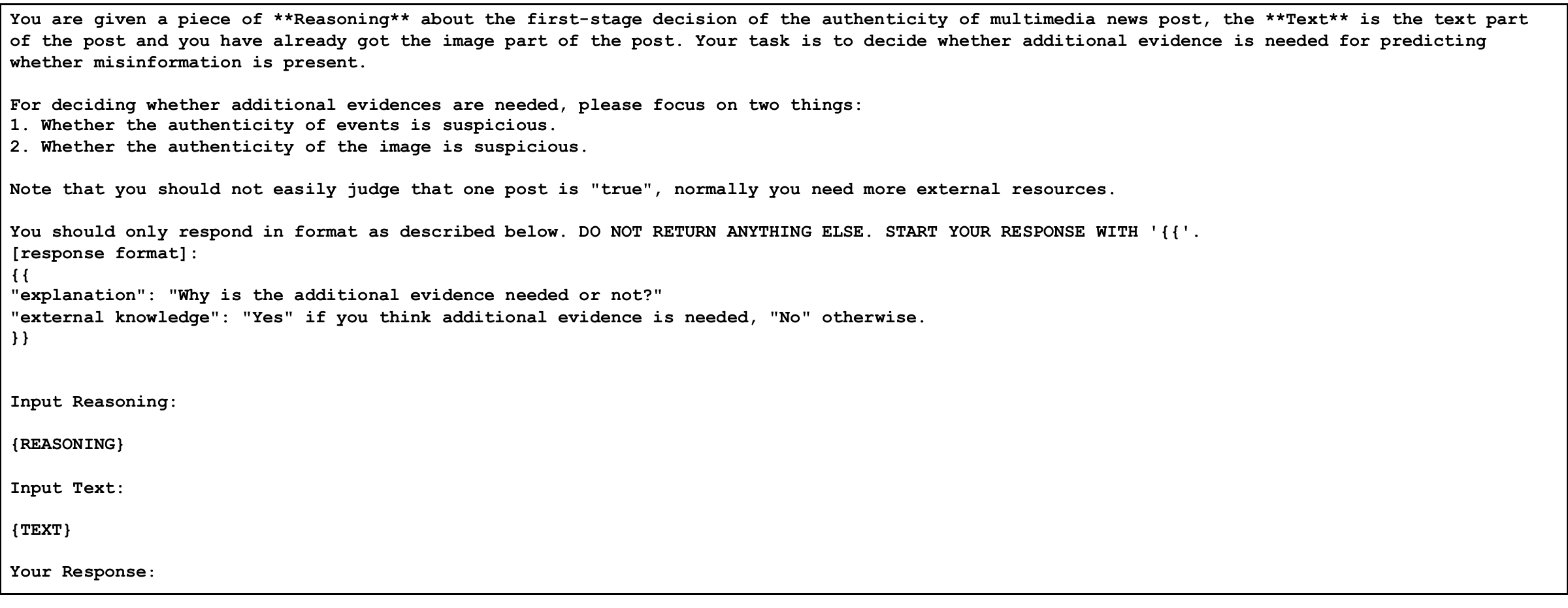}
    \caption{Instruction for External Knowledge}
    \label{fig: external knowledge}
\end{figure*}

\subsubsection{Necessity of External Knowledge} \label{sec: external knowledge prompt}
Based on the initial stage inference in Section \ref{sec:initial stage prompt}, LEMMA evaluates the necessity of incorporating external knowledge. This assessment is guided by rules specified in the prompt, which scrutinize the reasoning derived during the first stage. The decision to proceed to subsequent stages is contingent on whether the direct reasoning suggests the need for external verification to support or refute the findings. The detailed procedure of this evaluation is depicted in Figure \ref{fig: external knowledge}.

\subsubsection{Reasoning Aware Multi-query Generation}
\begin{figure*}[htb]
    \centering
    \includegraphics[width=0.95\textwidth]{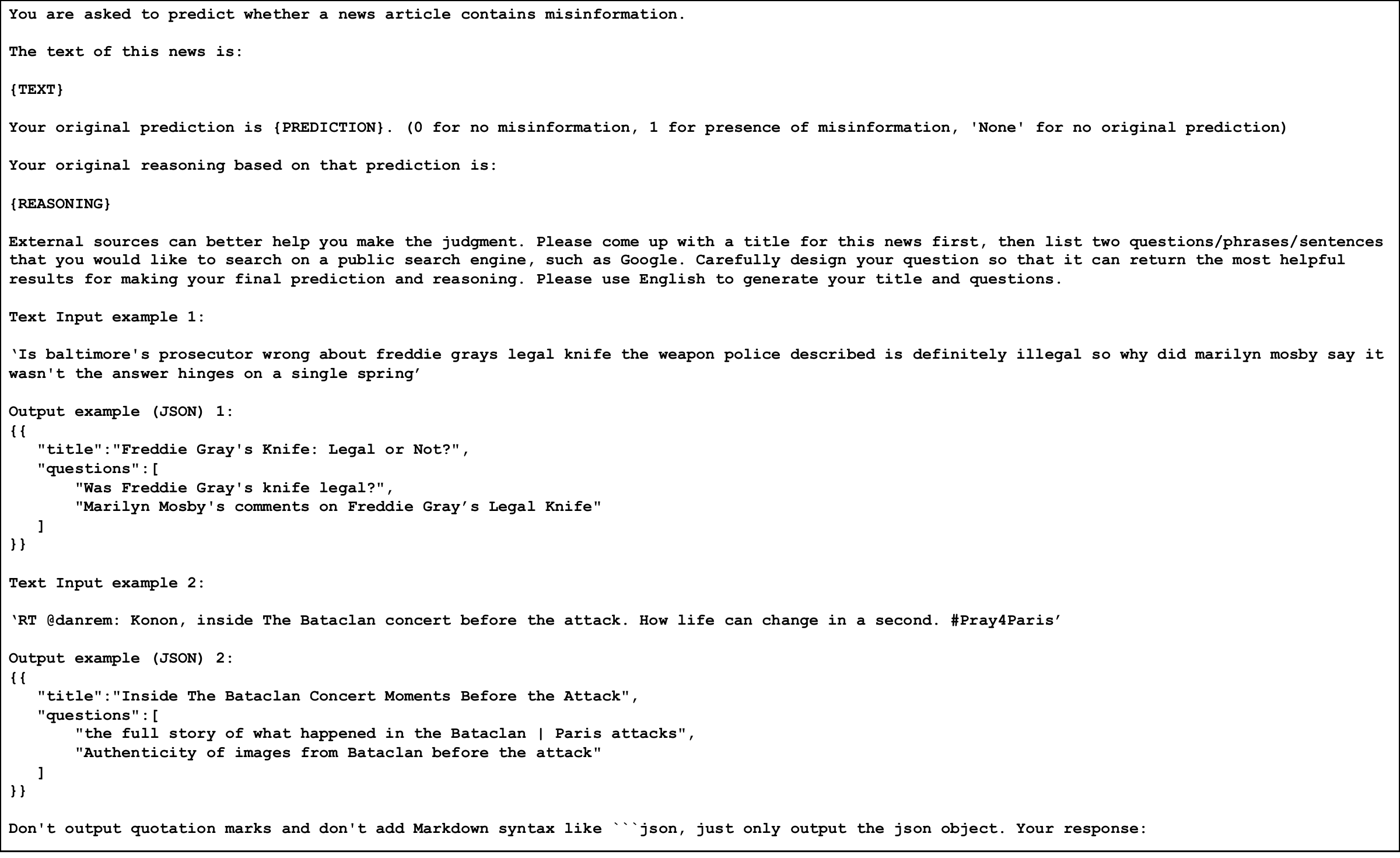}
    \caption{Instruction for Query Generation}
    \label{fig: Query}
\end{figure*}

At this stage, we input both the original image-text pair and the reasoning derived from \ref{sec:initial stage prompt} to generate the following queries: \textbf{1)} a title for the post. \textbf{2)} two questions related to contents that need to be verified. The design of the prompt is shown in Figure \ref{fig: Query}

\subsubsection{Topic Relevance Filter}
\begin{figure*}[htb]
    \centering
    \includegraphics[width=0.95\textwidth]{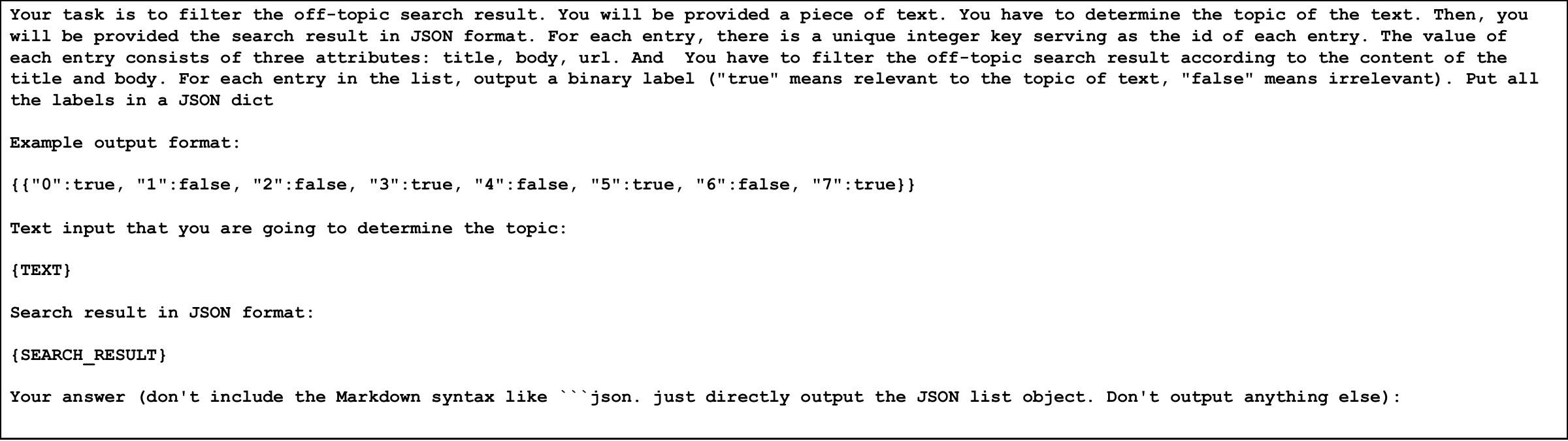}
    \caption{Instruction for Topic Relevance Filter}
    \label{fig: topic filter}
\end{figure*}

When we obtained the resources from the search engine (We use title and queries derived from \ref{sec: external knowledge prompt} to search resources), we use the following prompt to check whether each search result is related to the context based on the queries and title we derived in \ref{sec: external knowledge prompt}. Each resource will be labeled as True if found relevant, otherwise False. To ensure the efficiency, LVLM is asked to process a batch of  resources in one request, using JSON to manage the output format. The design of the prompt is shown in Figure \ref{fig: topic filter}

\subsubsection{Evidence Extraction}
\begin{figure*}[htb]
    \centering
    \includegraphics[width=0.95\textwidth]{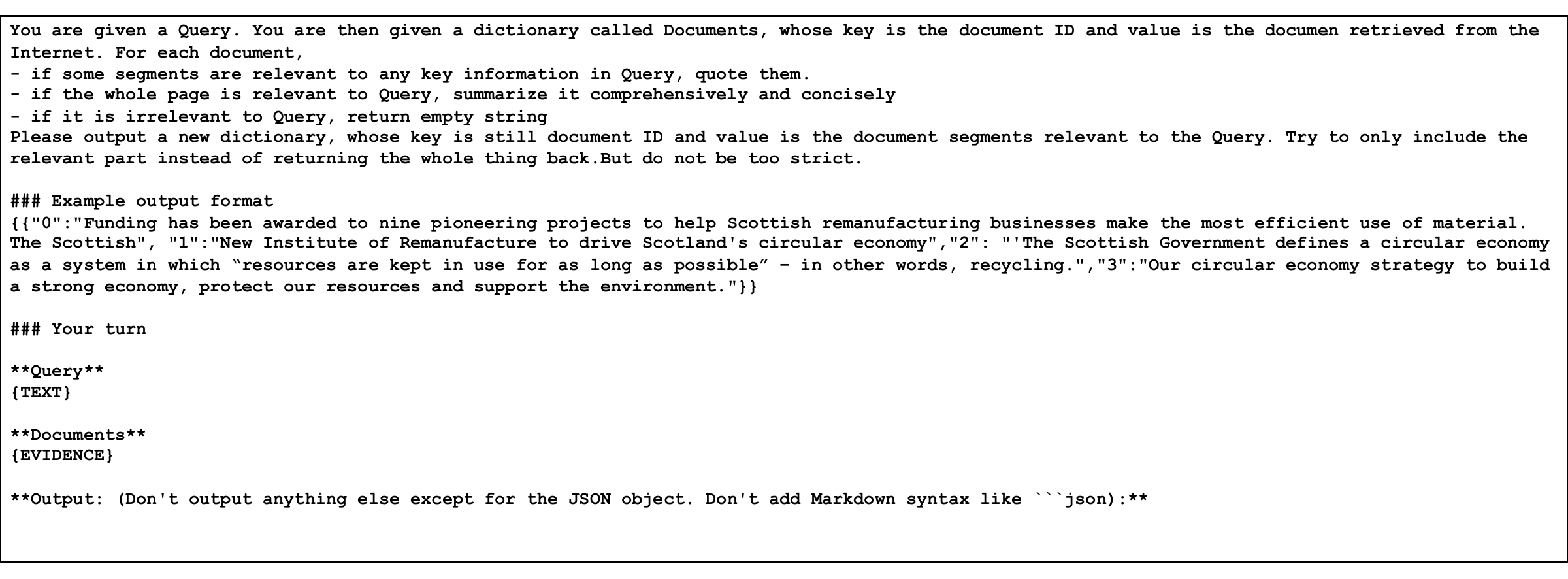}
    \caption{Instruction for Evidence Extraction}
    \label{fig: evidence}
\end{figure*}

This stage conducts evidence extraction by quoting or summarizing (if most of the post is relevant) the contents from remaining resources in the last stage. The model is asked to keep the extracted evidence concise, while avoiding excessive strictness. The design of the prompt is shown in Figure \ref{fig: evidence}

\subsubsection{Refined Prediction}
\begin{figure*}[htb]
    \centering
    \includegraphics[width=0.95\textwidth]{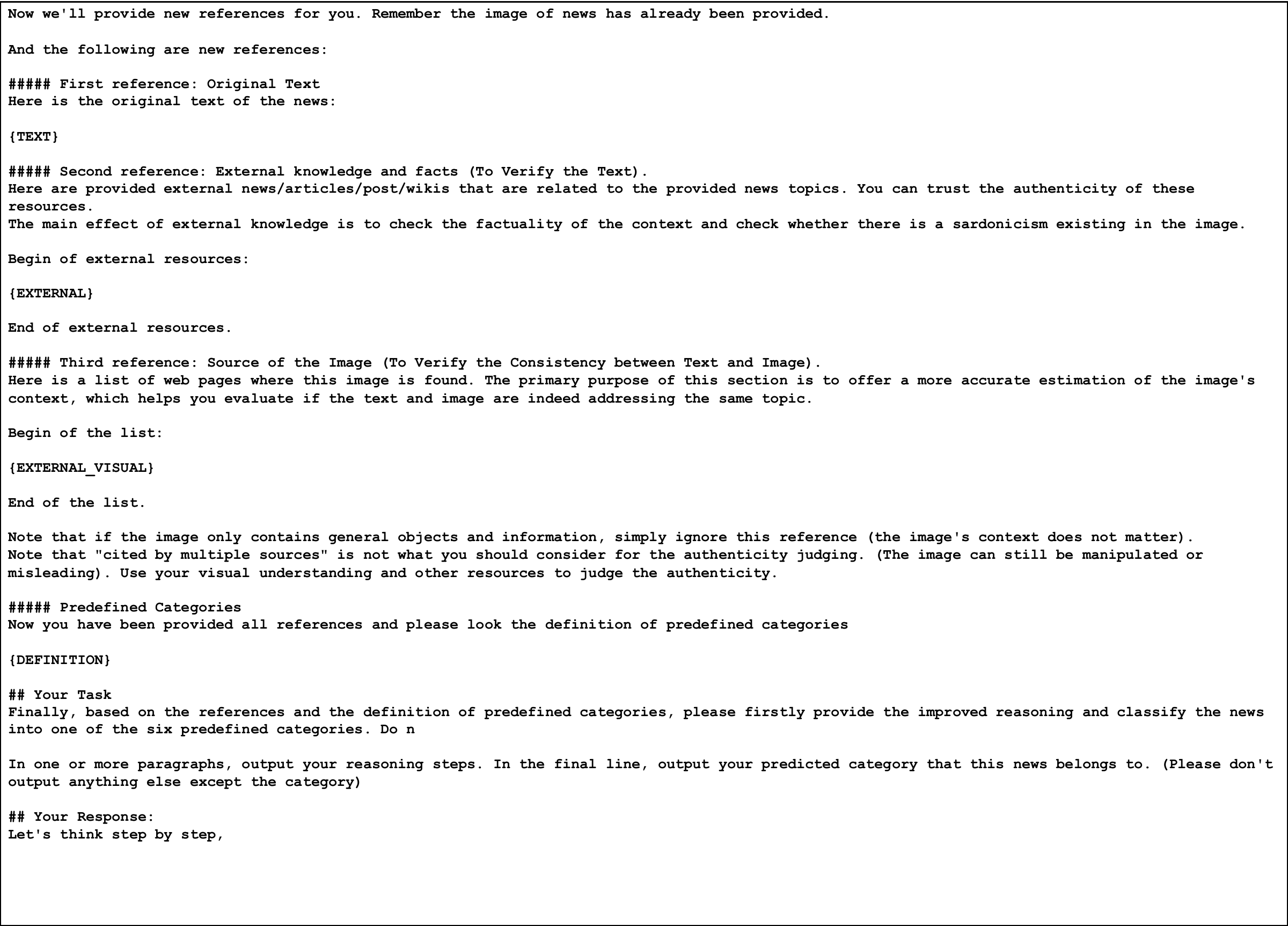}
    \caption{Instruction for Refined Prediction}
    \label{fig: refined}
\end{figure*}

Upon completing evidence extraction, the model reevaluates the image-text pair post, incorporating evidences retrieved from both text search and image search. Additionally, a fine-grained definition of misinformation (The detail is presented as Appendix \ref{sec: Appendix c}). is utilized for this reassessment. The design of the prompt is shown in Figure \ref{fig: refined}

\newpage
\section{Fine-grained definition of misinformation} \label{sec: Appendix c}
This section illustrates the fine-grained definition of misinformation which is used in Refined Prediction stage:
\begin{enumerate}
    \item \textbf{True}: True content is accurate in accordance with fact. Eight of the subreddits fall into this category, such as usnews and mildly interesting. The former consists of posts from various news sites. The latter encompasses real photos with accurate captions.
    \item \textbf{Satire/Parody}: This category includes content that presents true contemporary information in a satirical or humorous manner, often leading to its misinterpretation as false. Examples can be found on platforms like Reddit's \"The Onion,\" featuring headlines such as \"Man Lowers Carbon Footprint By Bringing Reusable Bags Every Time He Buys Gas.\", Satire that clearly identifies itself as such and is intended purely for entertainment or social commentary purposes should not be considered misinformation.
    \item \textbf{Misleading Content}: This category comprises information deliberately manipulated to deceive the audience.
    \item \textbf{False Connection}: This category encompasses instances where there is a disconnect between the information conveyed by an image and the essential details provided in the accompanying text. It may involve situations where the event depicted in the image does not align with or contradicts the narrative described in the text, leading to potential misinterpretation or misunderstanding.
    \item \textbf{Manipulated Content}: This category consists of content that has been intentionally altered through manual photo editing or other forms of manipulation. For instance, comments on platforms like the \"photoshopbattles\" subreddit often contain doctored versions of images submitted to the platform.
    \item \textbf{Unverified}: This category includes news or content for which the presence of misinformation cannot be definitively determined based solely on the available evidence. Additional evidence or verification may be required to confirm or refute the accuracy of the information. This category encompasses situations where there is insufficient information or conflicting sources to make a conclusive determination regarding the accuracy of the content. 
\end{enumerate}



\end{document}